\documentclass[sigconf]{acmart}

\usepackage{url}
\usepackage[english]{babel}
\usepackage[utf8x]{inputenc}
\usepackage[T1]{fontenc}

\usepackage{amsmath}
\usepackage{graphicx}
\usepackage[colorinlistoftodos]{todonotes}

\usepackage{amsthm}
\usepackage{amsfonts}
\usepackage{amsmath}

\usepackage{bm}
\usepackage{bbm}

\usepackage{subfig}

\usepackage{multirow}
\usepackage{threeparttable}
\usepackage{tabularx}
\usepackage{paralist}

\begin{document}

\copyrightyear{2019} 
\acmYear{2019} 
\setcopyright{acmlicensed} 
\acmConference[WSDM '19] {The Twelfth ACM International Conference on Web Search and Data Mining}{February 11--15, 2019}{Melbourne, VIC, Australia}
\acmPrice{15.00}
\acmDOI{10.1145/3289600.3290967}
 \acmISBN{978-1-4503-5940-5/19/02}

\title{SimGNN: A Neural Network Approach \\ to Fast Graph Similarity Computation
}

\author{
  Yunsheng Bai\textsuperscript{1}, Hao Ding\textsuperscript{2}, Song Bian\textsuperscript{3}, Ting Chen\textsuperscript{1}, Yizhou Sun\textsuperscript{1}, Wei Wang\textsuperscript{1}
    \\
  \textsuperscript{1}University of California, Los Angeles, CA, USA \\
  \textsuperscript{2}Purdue University, IN, USA \\
  \textsuperscript{3}Zhejiang University, China \\
  \texttt{yba@ucla.edu, ding209@purdue.edu, biansonghz@gmail.com, \{tingchen,yzsun,weiwang\}@cs.ucla.edu} \\
}

\begin{abstract}
Graph similarity search is among the most important graph-based applications, e.g. finding the chemical compounds that are most similar to a query compound. Graph similarity/distance computation, such as Graph Edit Distance (GED) and Maximum Common Subgraph (MCS), is the core operation of graph similarity search and many other applications, but very costly to compute in practice. Inspired by the recent success of neural network approaches to several graph applications,  such as node or graph classification, we propose a novel neural network based approach to address this classic yet challenging graph problem, aiming to alleviate the computational burden while preserving a good performance.  

The proposed approach, called SimGNN, combines two strategies. First, we design a learnable embedding function that maps every graph into an embedding vector, which provides a global summary of a graph. A novel attention mechanism is proposed to emphasize the important nodes with respect to a specific similarity metric. Second, we design a pairwise node comparison method to supplement the graph-level embeddings with fine-grained node-level information. Our model achieves better generalization on unseen graphs, and in the worst case runs in quadratic time with respect to the number of nodes in two graphs. Taking GED computation as an example, experimental results on three real graph datasets demonstrate the effectiveness and efficiency of our approach. Specifically, our model achieves smaller error rate and great time reduction compared against a series of baselines, including several approximation algorithms on GED computation, and many existing graph neural network based models. Our study suggests SimGNN provides a new direction for future research on graph similarity computation and graph similarity search\footnote{The code is publicly available at \url{https://github.com/yunshengb/SimGNN}.}.

\end{abstract}

\keywords{network embedding, neural networks, graph similarity computation, graph edit distance}

\maketitle
{\fontsize{8pt}{8pt} \selectfont
\textbf{ACM Reference Format:}\\
Yunsheng Bai, Hao Ding, Song Bian, Ting Chen, Yizhou Sun, Wei Wang. 2019. SimGNN: A Neural Network Approach to Fast Graph Similarity Computation. In The Twelfth ACM International Conference on Web Search and Data Mining (WSDM’19), February 11--15, 2019, Melbourne, VIC, Australia. ACM, New York, NY, USA, 9 pages. https://doi.org/10.1145/3289600.3290967 
} 
\section{Introduction}
\label{sec-intro}

Graphs are ubiquitous nowadays and have a wide range of applications in bioinformatics, chemistry, recommender systems, social network study, program static analysis, etc. Among these, one of the fundamental problems is to retrieve a set of similar graphs from a database given a user query. Different graph similarity/distance metrics are defined, such as Graph Edit Distance (GED)~\cite{bunke1983distance}, Maximum Common Subgraph (MCS)~\cite{bunke1998graph}, etc. However, the core operation, namely computing the GED or MCS between two graphs, is known to be NP-complete~\cite{zeng2009comparing,bunke1998graph}. For GED, even the state-of-the-art algorithms cannot reliably compute the exact GED within reasonable time between graphs with more than 16 nodes~\cite{blumenthal2018exact}.

\begin{figure}
\centering
\includegraphics[scale=0.28]{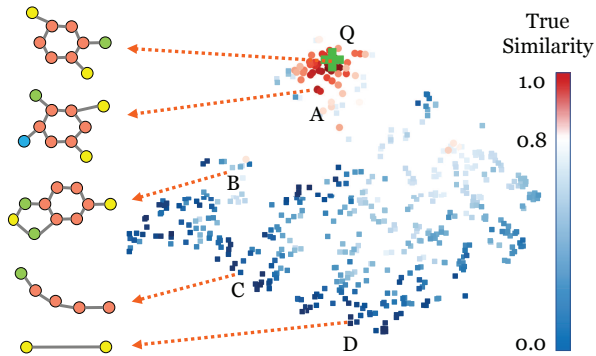}
\vspace{-0.1in}
\caption{Illustration of similarity-preserving graph embedding. Each graph is mapped into an embedding vector (denoted as a dot in the plot), which preserves their similarity between each other in terms of a specific graph similarity metric. The green ``+'' sign denotes the embedding of an example query graph. Colors of dots indicate how similar a graph is to the query based on the ground truth (from red to blue, meaning from the most similar to the least similar).
}
\label{fig:intro}
\vspace*{-5mm}
\end{figure}

Given the huge importance yet great difficulty of computing the exact graph distances, there have been two broad categories of methods to address the problem of graph similarity search. The first category of remedies is the pruning-verification framework~\cite{zeng2009comparing,zhao2013partition,liang2017similarity}, under which the total amount of exact graph similarity computations for a query can be reduced to a tractable degree, via a series of database indexing techniques and pruning strategies. However, the fundamental problem of the exponential time complexity of exact graph similarity computation~\cite{neuhaus2006fast} 
 remains. The second category tries to reduce the cost of graph similarity computation directly. Instead of calculating the exact similarity metric, these methods find approximate values in a fast and heuristic way~\cite{neuhaus2006fast,riesen2009approximate,fankhauser2011speeding,bougleux2017graph,daller2018approximate}. However, these methods usually require rather complicated design and implementation based on discrete optimization or combinatorial search. The time complexity is usually still polynomial or even sub-exponential in the number of nodes in the graphs, such as A*-Beamsearch (Beam)~\cite{neuhaus2006fast}, Hungarian~\cite{riesen2009approximate}, VJ~\cite{fankhauser2011speeding}, etc. 

In this paper, we propose a novel approach to speed-up the graph similarity computation, with the same purpose as the second category of methods mentioned previously. However, instead of directly computing the approximate similarities using combinatorial search, our solution turns it into a \emph{learning} problem. More specifically, we design a neural network-based function that maps a pair of graphs into a similarity score.  At the training stage, the parameters involved in this function will be learned by minimizing the difference between the predicted similarity scores and the ground truth, where each training data point is a pair of graphs together with their true similarity score. At the test stage, by feeding the learned function with any pair of graphs, we can obtain a predicted similarity score. We name such approach as \emph{\textbf{SimGNN}}, i.e., \emph{\underline{Sim}}ilarity Computation via \emph{\underline{G}}raph \emph{\underline{N}}eural \emph{\underline{N}}etworks. 


SimGNN enjoys the key advantage of efficiency due to the nature of neural network computation. As for effectiveness, however, we need to carefully design the neural network architecture to satisfy the following three properties:
\begin{enumerate}\item\textit{\textbf{Representation-invariant}}. The same graph can be represented by different adjacency matrices by permuting the order of nodes. The computed similarity score should be invariant to such changes. \item\textit{\textbf{Inductive}}. The similarity computation should generalize to unseen graphs, i.e. compute the similarity score for graphs outside the training graph pairs. \item \textit{\textbf{Learnable}}. The model should be adaptive to any similarity metric, by adjusting its parameters through training. 
\end{enumerate}
To achieve these goals, we propose the following two strategies. First, we design a learnable embedding function that maps every graph into an embedding vector, which provides a global summary of a graph through aggregating node-level embeddings. We propose a novel attention mechanism to select the important nodes out of an entire graph with respect to a specific similarity metric. This graph-level embedding can already largely preserve the similarity between graphs. For example, as illustrated in Fig.~\ref{fig:intro}, Graph $A$ is very similar to Graph $Q$ according to the ground truth similarity, which is reflected by the embedding as its embedding is close to $Q$ in the embedding space. Also, such embedding-based similarity computation is very fast. Second, we design a pairwise node comparison method to supplement the graph-level embeddings with fine-grained node-level information. As one fixed-length embedding per graph may be too coarse, we further compute the pairwise similarity scores between nodes from the two graphs, from which the histogram features are extracted and combined with the graph-level information to boost the performance of our model. This results in the quadratic amount of operations in terms of graph size, which, however,  is still among the most efficient methods for graph similarity computation.

We conduct our experiments on GED computation, which is one of the most popular graph similarity/distance metrics. To demonstrate the effectiveness and efficiency of our approach, we conduct experiments on three real graph datasets. Compared with the baselines, which include several approximate GED computation algorithms, and many graph neural network based methods, our model achieves smaller error and great time reduction. It is worth mentioning that, our Strategy 1 already demonstrates superb performances compared with existing solutions. When running time is a major concern, we can drop the more time-consuming Strategy 2 for trade-off.

Our contributions can be summarized as follows:
\vspace{-\topsep}
\begin{itemize}
\item We address the challenging while classic problem of graph similarity computation by considering it as a learning problem, and propose a neural network based approach, called SimGNN, as the solution.
\item Two novel strategies are proposed. First, we propose an efficient and effective attention mechanism to select the most relevant parts of a graph to generate a graph-level embedding, which preserves the similarity between graphs. Second, we propose a pairwise node comparison method to supplement the graph-level embeddings for more effective modeling of the similarity between two graphs.
\item We conduct extensive experiments on a very popular graph similarity/distance metric, GED,  based on three real network datasets to demonstrate the effectiveness and efficiency of the proposed approach.
\end{itemize}

The rest of this paper is organized as follows. We introduce the preliminaries of our work in Section~\ref{sec-prelim}, describe our model in Section~\ref{sec-model}, present experimental results in Section~\ref{sec-exp}, discuss related work in Section~\ref{sec-related}, and point out future directions in Section~\ref{sec-future}. A conclusion is provided in Section~\ref{sec-conc}.

\section{Preliminaries}
\label{sec-prelim}
\subsection{Graph Edit Distance (GED)}
In order to demonstrate the effectiveness and efficiency of SimGNN, we choose one of the most popular graph similarity/distance metric, Graph Edit Distance (GED), as a case study.  GED has been widely used in many applications, such as graph similarity search~\cite{zeng2009comparing,wang2012efficient,zheng2013graph,zhao2013partition,liang2017similarity}, graph classification~\cite{riesen2008iam,riesen2009approximate}, handwriting recognition~\cite{fischer2013fast}, image indexing~\cite{xiao2008hmm}, etc. 

\begin{figure}
\centering
\includegraphics[scale=0.27]{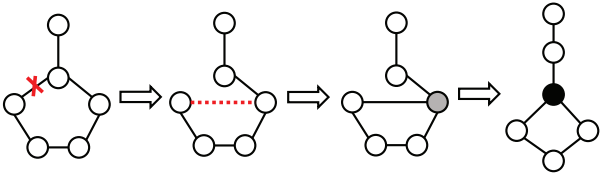}
\vspace{-0.1in}
\caption{The GED between the graph to the left and the graph to the right is 3, as the transformation needs 3 edit operations: (1) an edge deletion, (2) an edge insertion, and (3) a node relabeling.}
\vspace*{-6mm}
\label{fig:ged}
\end{figure}

Formally, the edit distance between $\mathcal{G}_1$ and $\mathcal{G}_2$, denoted by $\mathrm{GED} (\mathcal{G}_1,\mathcal{G}_2)$, is the number of edit operations in the optimal alignments that transform $\mathcal{G}_1$ into $\mathcal{G}_2$, where an edit operation on a graph $\mathcal{G}$ is an insertion or deletion of a vertex/edge or relabelling of a vertex \footnote{Although other variants of GED exist~\cite{riesen2013novel}, we adopt this basic version.}.
Intuitively, if two graphs are identical (isomorphic), their GED is 0. Fig.~\ref{fig:ged} shows an example of GED between two simple graphs. 

Once the distance between two graphs is calculated, we transform it to a similarity score ranging between 0 and 1. More details about the transformation function can be found in Section~\ref{subsec-data-preproc}.

\subsection{Graph Convolutional Networks (GCN)}
Both strategies in SimGNN require node embedding computation. In Strategy 1, to compute graph-level embedding, it aggregates node-level embeddings using attention; and in Strategy 2, pairwise node comparison for two graphs is computed based on node-level embeddings as well. 

Among many existing node embedding algorithms, we choose to use Graph Convolutional Networks (GCN)~\cite{kipf2016semi}, as it is graph \emph{representation-invariant}, as long as the initialization is carefully designed. It is also \emph{inductive}, since for any unseen graph, we can always compute the node embedding following the GCN operation. GCN now is among the most popular models for node embeddings, and belong to the family of neighbor aggregation based methods. Its core operation, graph convolution, operates on the representation of a node, which is denoted as $\bm{u_n} \in \mathbb{R}^{D}$, and is defined as follows:

\begin{equation} \label{eq:gcn} \mathrm{conv}(\bm{u_n}) = f_1( \sum_{m \in \mathcal{N}(n)} \frac{1}{\sqrt[]{d_n d_m}} \bm{u_m} \bm{W_{1}}^{(l)} + \bm{b_{1}}^{(l)})\end{equation} 
where $\mathcal{N}(n)$ is the set of the first-order neighbors of node $n$ plus $n$ itself, $d_n$ is the degree of node $n$ plus 1, $\bm{W_{1}}^{(l)} \in \mathbb{R}^{D^{l} \times D^{l+1}}$ is the weight matrix associated with the $l$-th GCN layer, $\bm{b_{1}}^{(l)} \in \mathbb{R}^{D^{l+1}}$ is the bias, and $f_1(\cdot)$ is an activation function such as $\mathrm{ReLU}(x)=\mathrm{max}(0,x)$. Intuitively, the graph convolution operation aggregates the features from the first-order neighbors of the node. 

\section{The Proposed Approach: SimGNN}
\label{sec-model}
\begin{figure*}
\centering
\includegraphics[scale=0.22]{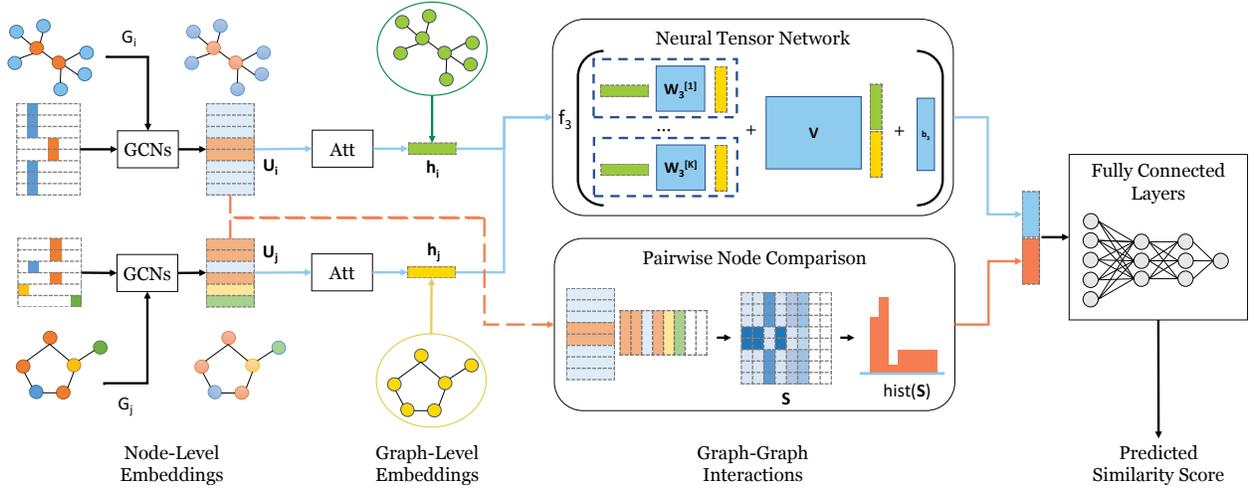}
\vspace*{-3mm}
\caption{An overview illustration of SimGNN. The blue arrows denote the data flow for Strategy 1, which is based on graph-level embeddings. The red arrows denote the data flow for Strategy 2, which is based on pairwise node comparison.}
\label{fig:model}
\vspace*{-4mm}
\end{figure*}

Now we introduce our proposed approach SimGNN in detail, which is an end-to-end neural network based approach that attempts to learn a function to map a pair of graphs into a similarity score.
An overview of SimGNN is illustrated in Fig.~\ref{fig:model}. First, our model transforms the node of each graph into a vector, encoding the features and structural properties around each node. Then, two strategies are proposed to model the similarity between the two graphs, one based on the interaction between two graph-level embeddings, the other based on comparing two sets of node-level embeddings. Finally, two strategies are combined together to feed into a fully connected neural network to get the final similarity score. The rest of the section details these two strategies. 

\subsection{Strategy One: Graph-Level Embedding Interaction}

This strategy is based on the assumption that a good graph-level embedding can encode the structural and feature information of a graph, and by interacting the two graph-level embeddings, the similarity between two graphs can be predicted. It involves the following stages: (1) the node embedding stage, which transforms each node of a graph into a vector, encoding its features and structural properties; (2) the graph embedding stage, which produces one embedding for each graph by attention-based aggregation of node embeddings generated in the previous stage; 
(3) the graph-graph interaction stage, which receives two graph-level embeddings and returns the interaction scores representing the graph-graph similarity; and (4) the final graph similarity score computation stage, which further reduces the interaction scores into one final similarity score. It will be compared against the ground-truth similarity score to update parameters involved in the 4 stages.

\subsubsection{Stage I: Node Embedding}
\label{subsec-node-emb}
Among the existing state-of-the-art approaches, we adopt GCN, a neighbor aggregation based method, because it learns an aggregation function (Eq.~\ref{eq:gcn}) 
that are representation-invariant and can be applied to unseen nodes. In Fig.~\ref{fig:model}, different colors represent different node types, and the original node representations are one-hot encoded. Notice that the one-hot encoding is based on node types (e.g., all the nodes with carbon type share the same one-hot encoding vector), so even if the node ids are permuted, the aggregation results would be the same. For graphs with unlabeled nodes, we treat every node to have the same label, resulting in the same constant number as the initialize representation. 
After multiple layers of GCNs (e.g., 3 layers in our experiment),  the node embeddings are ready to be fed into the Attention module (Att), which is described as follows.

\subsubsection{Stage II: Graph Embedding: Global Context-Aware Attention.}
\label{subsec-graph-emb}
To generate one embedding per graph using a set of node embeddings, one could perform an unweighted average of node embeddings, or a weighted sum where a weight associated with a node is determined by its degree. 
However, which nodes are more important and should receive more weights is dependent on the specific similarity metric. Thus, we propose the following attention mechanism to let the model learn weights guided by the specific similarity metric.

Denote the input node embeddings as $\bm{U} \in \mathbb{R}^{N \times D}$, where the $n$-th row, $\bm{u_n} \in \mathbb{R}^{D}$ is the embedding of node $n$. First, a global graph context $\bm{c} \in \mathbb{R}^{D}$ 
is computed, which is a simple average of node embeddings followed by a nonlinear transformation: $\bm{c} = \mathrm{tanh}(\frac{1}{N} \bm{W_2} \sum_{n=1}^{N} \bm{u_n} )$, where $\bm{W_2} \in \mathbb{R}^{D \times D}$ is a learnable weight matrix. The context $\bm{c}$ provides the global structural and feature information of the graph that is adaptive to the given similarity metric, via learning the weight matrix. Based on $\bm{c}$, we can compute one attention weight for each node.


For node $n$, to make its attention $a_n$ aware of the global context, we take the inner product between $\bm{c}$ and its node embedding. The intuition is that, nodes similar to the global context should receive higher attention weights. A sigmoid function $\sigma(x)=\frac{1}{1+\exp{(-x)}}$ is applied to the result to ensure the attention weights is in the range $(0,1)$. We do not normalize the weights into length 1, since it is desirable to let the embedding norm reflect the graph size, which is essential for the task of graph similarity computation. Finally, the graph embedding $\bm{h} \in \mathbb{R}^{D}$ is the weighted sum of node embeddings, $\bm{h} = \sum_{n=1}^{N} a_n \bm{u_n}$. The following equation summarizes the proposed node attentive mechanism:
\begin{equation}\bm{h} = \sum_{n=1}^{N} f_2(\bm{u_n^{T}} \bm{c}) \bm{u_n} = \sum_{n=1}^{N} f_2(\bm{u_n^{T}} \mathrm{tanh}(\frac{1}{N} \bm{W_2} \sum_{m=1}^{N} \bm{u_m}) ) \bm{u_n}\end{equation}
where $f_2(\cdot)$ is the sigmoid function $\sigma(\cdot)$.

\subsubsection{Stage III: Graph-Graph Interaction: Neural Tensor Network}
Given the graph-level embeddings of two graphs produced by the previous stage, a simple way to model their relation is to take the inner product of the two, $\bm{h_i} \in \mathbb{R}^{D}$, $\bm{h_j} \in \mathbb{R}^{D}$. However, as discussed in \cite{socher2013reasoning}, such simple usage of data representations often lead to insufficient or weak interaction between the two. Following \cite{socher2013reasoning}, we use Neural Tensor Networks (NTN) to model the relation between two graph-level embeddings:
\begin{equation}g(\bm{h_i}, \bm{h_j}) = f_3(\bm{h_{i}^{T}} \bm{W_{3}^{[1:K]}} \bm{h_{j}} + \bm{V} \bigl[\begin{smallmatrix} \bm{h_{i}} \\ \bm{h_{j}} \end{smallmatrix}\bigr] + \bm{b_3})\end{equation}
where $\bm{W_{3}^{[1:K]}} \in \mathbb{R}^{D \times D \times K}$ is a weight tensor, $[]$ denotes the concatenation operation, $\bm{V} \in \mathbb{R}^{K \times 2D}$ is a weight vector, $\bm{b_3} \in \mathbb{R}^{K}$ is a bias vector, and $f_3(\cdot)$ is an activation function. $K$ is a hyperparameter controlling the number of interaction (similarity) scores produced by the model for each graph embedding pair. 

\subsubsection{Stage IV: Graph Similarity Score Computation}
\label{subsec-gsp}

After obtaining a list of similarity scores, we apply a standard multi-layer fully connected neural network to gradually reduce the dimension of the similarity score vector. In the end, one score, $\hat{s_{ij}} \in \mathbb{R}$, is predicted, and it is compared against the ground-truth similarity score using the following mean squared error loss function:

\begin{equation} \mathcal{L} = \frac{1}{|\mathcal{D}|}\sum_{(i,j) \in \mathcal{D}} (\hat{s_{ij}} - s (\mathcal{G}_i,\mathcal{G}_j))^{2}\end{equation}
where $\mathcal{D}$ is the set of training graph pairs, and $s (\mathcal{G}_i,\mathcal{G}_j)$ is the ground-truth similarity between $\mathcal{G}_i$ and $\mathcal{G}_j$.

\subsubsection{Limitations of Strategy One}
As mentioned in Section~\ref{sec-intro}, the node-level information such as the node feature distribution and graph size may be lost by the graph-level embedding. In many cases, the differences between two graphs lie in small substructures and are hard to be reflected by the graph level embedding. 
An analogy is that, in Natural Language Processing, the performance of sentence matching based on one embedding per sentence can be further enhanced through using fine-grained word-level information~\cite{hu2014convolutional,he2016pairwise}. This leads to our second strategy.


\subsection{Strategy Two: Pairwise Node Comparison}
\label{subsec-pnc}
To overcome the limitations mentioned previously, we consider bypassing the NTN module, and using the node-level embeddings directly. As illustrated in the bottom data flow of Fig.~\ref{fig:model}, if $\mathcal{G}_i$ has $N_i$ nodes and $\mathcal{G}_j$ has $N_j$ nodes, there would be $N_i N_j$ pairwise interaction scores, obtained by $\bm{S} = \sigma(\bm{U_i} \bm{U_j^{T}})$, where $\bm{U_i} \in \mathbb{R}^{N_i \times D}$ and $\bm{U_j} \in \mathbb{R}^{N_j \times D}$ are the node embeddings of $\mathcal{G}_i$ and $\mathcal{G}_j$, respectively. Since the node-level embeddings are not normalized, the sigmoid function is applied to ensure the similarities scores are in the range of $(0,1)$. For two graphs of different sizes, to emphasize their size difference, we pad fake nodes to the smaller graph. As shown in Fig.~\ref{fig:model}, two fake nodes with zero embedding are padded to the bottom graph, resulting in two extra columns with zeros in $\bm{S}$.

Denote $N=\mathrm{max}(N_1, N_2)$. The pairwise node similarity matrix $\bm{S} \in \mathbb{R}^{N\times N}$ is a useful source of information, since it encodes fine-grained pairwise node similarity scores. One simple way to utilize $\bm{S}$ is to vectorize it: $\mathrm{vec}(\bm{S}) \in \mathbb{R}^{N^2}$, and feed it into the fully connected layers. However, there is usually no natural ordering between graph nodes. Different initial node ordering of the same graph would lead to different $\bm{S}$ and $\mathrm{vec}(\bm{S})$.

To ensure the model is invariant to the graph representations as mentioned in Section~\ref{sec-intro}, one could preprocess the graph by applying some node ordering scheme~\cite{niepert2016learning}, but we consider a much more efficient and natural way to utilize $\bm{S}$. We extract its histogram features: $\mathrm{hist}(\bm{S}) \in \mathbb{R}^{B}$, where $B$ is a hyperparameter that controls the number of bins in the histogram. In the case of Fig.~\ref{fig:model}, seven bins are used for the histogram. The histogram feature vector is normalized and concatenated with the graph-level interaction scores $g(\bm{h_i}, \bm{h_j})$, and fed to the fully connected layers to obtain a final similarity score for the graph pair.

\begin{table}
\caption{Statistics of datasets.}
\small
\begin{tabular}
{|c|ccc|} \hline
\textbf{Dataset} & \textbf{Graph Meaning} & \textbf{\#Graphs} & \textbf{\#Pairs} \\ \hline
\textbf{AIDS} & {Chemical Compounds} & 700 & 490K \\ \hline
\textbf{LINUX} & {Program Dependency Graphs} & 1000 & 1M  \\ \hline
\textbf{IMDB} & {Actor/Actress Ego-Networks} & 1500 & 2.25M\\ \hline
\end{tabular}
\centering
\label{dataset_summary}
\vspace*{-4mm}
\end{table}

It is important to note that the histogram features alone are not enough to train the model, since the histogram is not a continuous differential function and does not support backpropagation. In fact, we rely on Strategy 1 as the primary strategy to update the model weights, and use Strategy 2 to supplement the graph-level features, which brings extra performance gain to our model. 

To sum up, we combine the coarse global comparison information captured by Strategy 1, and the fine-grained node-level comparison information captured by Strategy 2, to provide a thorough view of the graph comparison to the model.

\subsection{Time Complexity Analysis}
\label{subsec-time-comp-analysis}
Once SimGNN has been trained, it can be used to compute a similarity score for any pair of graphs. The time complexity involves two parts: (1) the node-level and global-level embedding computation stages, which needs to be computed once for each graph; and (2) the similarity score computation stage, which needs to be computed for every pair of graphs.

\noindent\textbf{The node-level and global-level embedding computation stages.} The time complexity associated with the generation of node-level and graph-level embeddings is $O(E)$~\cite{kipf2016semi}, where $E$ is the number of edges of the graph. Notice that the graph-level embeddings can be pre-computed and stored, and in the setting of graph similarity search, the unseen query graph only needs to be processed once to obtain its graph-level embedding. 

\noindent\textbf{The similarity score computation stage.}
The time complexity for Strategy 1 is $O(D^2 K)$,  where $D$ is the dimension of the graph level embedding, and $K$ is the feature map dimension of the NTN. 
The time complexity for our Strategy 2 is $O(D N^2)$, where $N$ is the number of nodes in the larger graph. This can potentially be reduced by node sampling to construct the similarity matrix $S$. Moreover, the matrix multiplication $\bm{S} = \sigma(\bm{U_1} \bm{U_2^{T}})$ can be greatly accelerated with GPUs. Our experimental results in Section~\ref{subsec-efficiency} verify that there is no significant runtime increase when the second strategy is used.

In conclusion, among the two strategies we have proposed: Strategy 1 is the primary strategy, which is efficient but solely based on coarse graph level embeddings; and Strategy 2 is auxiliary, which includes fine-grained node-level information but is more time-consuming. In the worst case, the model runs in quadratic time with respect to the number of nodes, which is among the state-of-the-art algorithms for approximate graph distance computation.


\section{Experiments} 
\label{sec-exp}

\subsection{Datasets}
\begin{figure*}
    \centering
    \subfloat[Node label distribution of AIDS.]{{\includegraphics[scale=0.32]{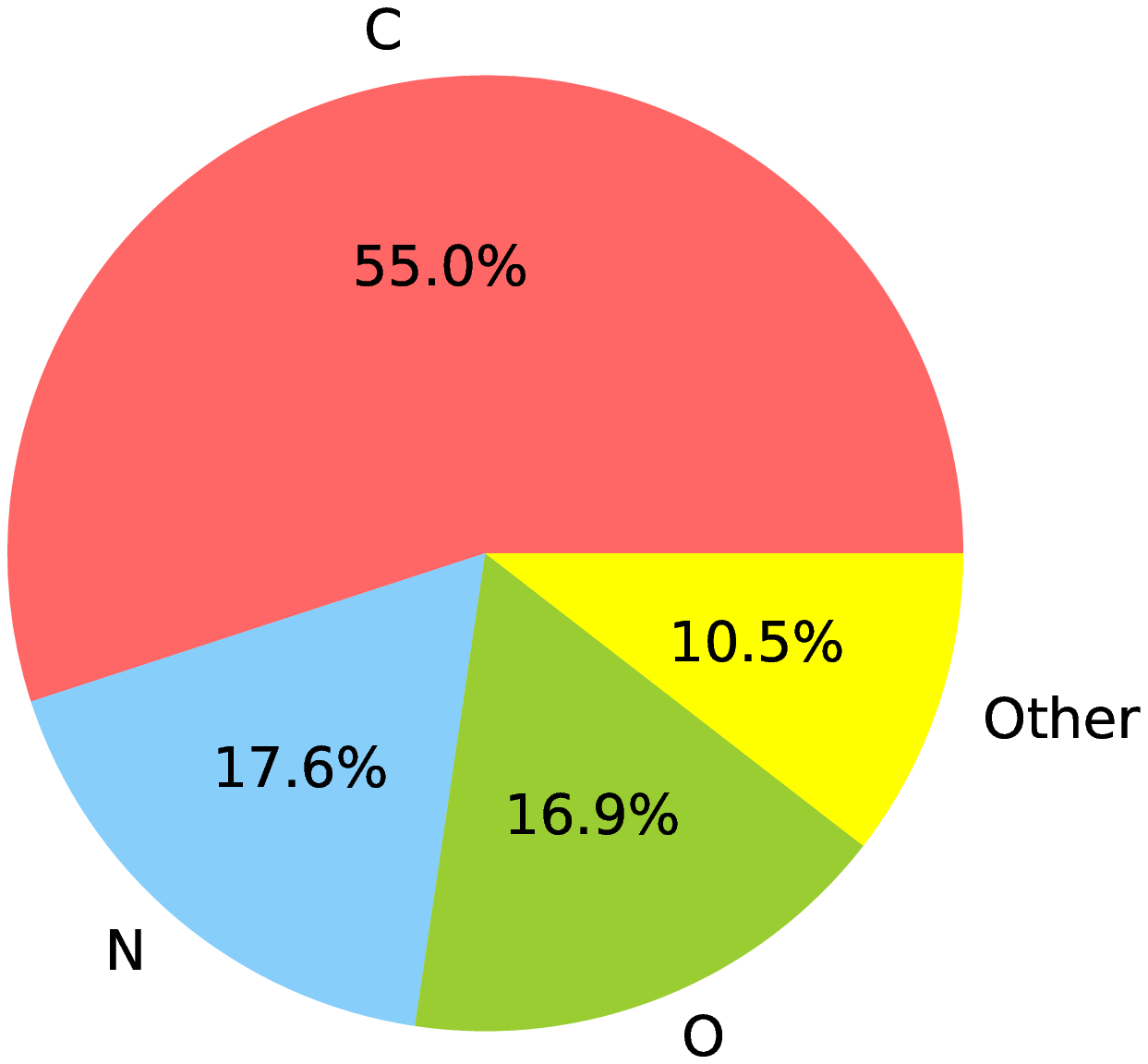} } \label{fig:aids_pie}}
    \subfloat[Distribution of graph sizes.]{{\includegraphics[scale=0.28]{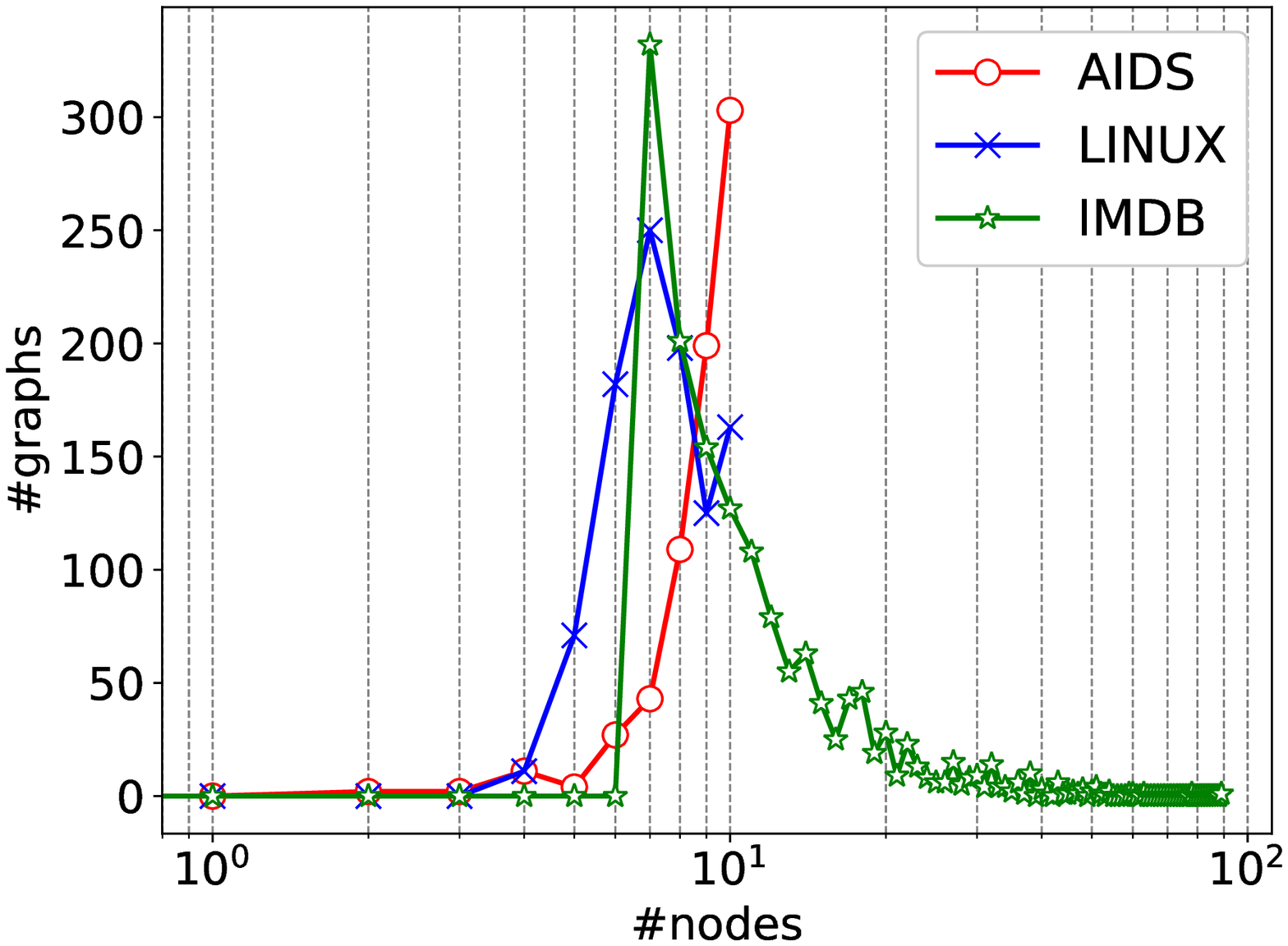} }}
    \subfloat[Distribution of GEDs of the training pairs.]{{\includegraphics[scale=0.28]{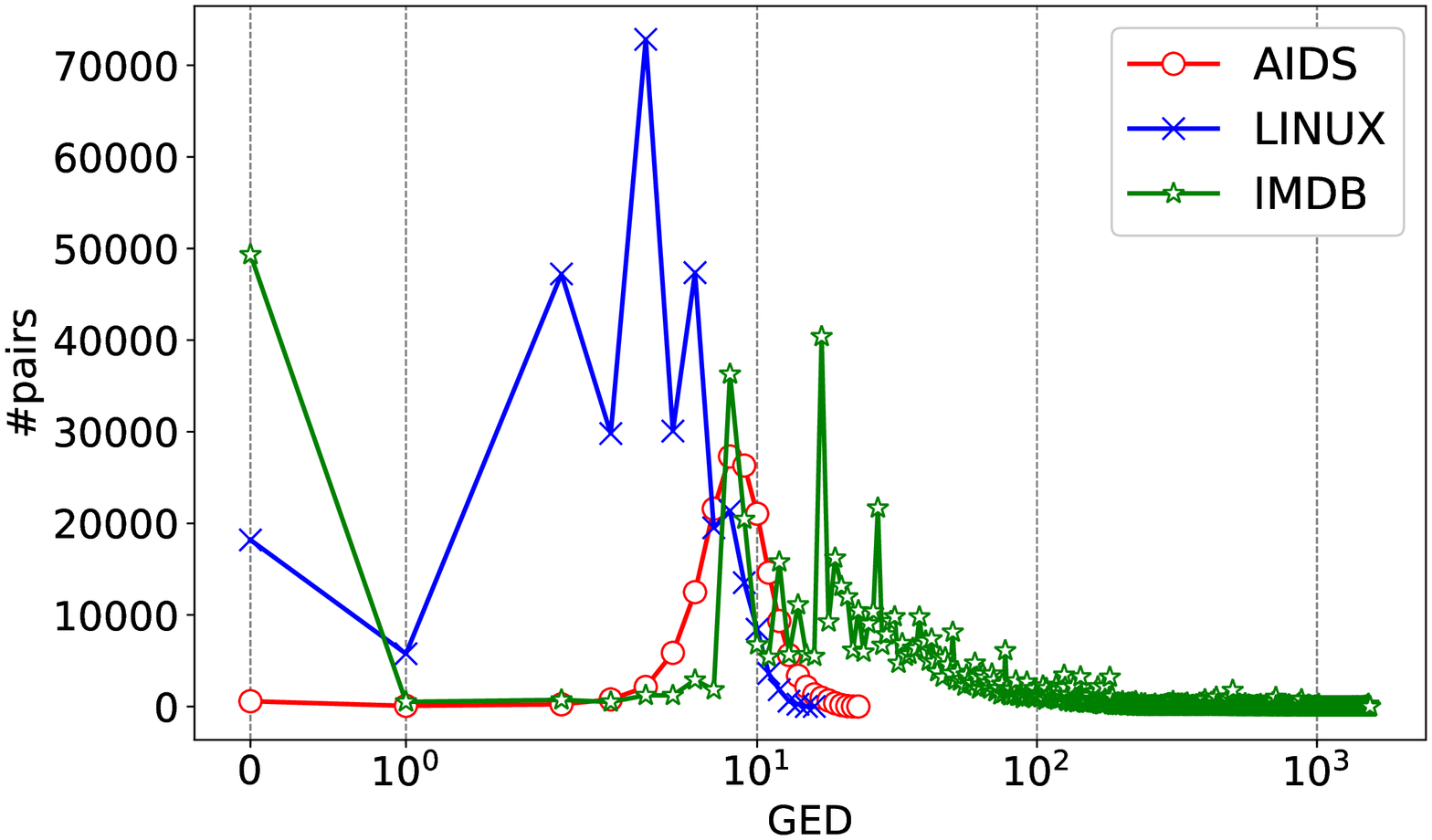} } \label{fig:ged_distribution}}
    \vspace{-0.1in}
    \caption{Some statistics of the datasets.}
    \vspace{-0.1in}
    \label{fig:dataset_vis}
\end{figure*}

Three real-world graph datasets are used for the experiments. A concise summary and detailed visualizations can be found in Table~\ref{dataset_summary} and Fig.~\ref{fig:dataset_vis}, respectively.

\textit{AIDS}. AIDS is a collection of antivirus screen chemical compounds from the Developmental Therapeutics Program at NCI/NIH 7~\footnote{\url{https://wiki.nci.nih.gov/display/NCIDTPdata/AIDS+Antiviral+Screen+Data}}, and has been used in several existing works on graph similarity search~\cite{zeng2009comparing,wang2012efficient,zheng2013graph,zhao2013partition,liang2017similarity}. It contains 42,687 chemical compound structures with Hydrogen atoms omitted. We select 700 graphs, each of which has 10 or less than 10 nodes. Each node is labeled with one of 29 types, as illustrated in Fig.~\ref{fig:aids_pie}. 

\textit{LINUX}. The LINUX dataset was originally introduced in \cite{wang2012efficient}. It consists of 48,747 Program Dependence Graphs (PDG) generated from the Linux kernel. Each graph represents a function, where a node represents one statement and an edge represents the dependency between the two statements. We randomly select 1000 graphs of equal or less than 10 nodes each. The nodes are unlabeled.

\textit{IMDB}. The IMDB dataset~\cite{yanardag2015deep} (named ``IMDB-MULTI'') consists of 1500 ego-networks of movie actors/actresses, where there is an edge if the two people appear in the same movie. To test the scalability and efficiency of our proposed approach, we use the full dataset without any selection. The nodes are unlabeled.


\begin{table}
\caption{Statistics of datasets.}
\small
\vspace*{-4mm}
\begin{tabular}
{|c|ccc|} \hline
\textbf{Dataset} & \textbf{Graph Meaning} & \textbf{\#Graphs} & \textbf{\#Pairs} \\ \hline
\textbf{AIDS} & {Chemical Compounds} & 700 & 490K \\ \hline
\textbf{LINUX} & {Program Dependency Graphs} & 1000 & 1M  \\ \hline
\textbf{IMDB} & {Actor/Actress Ego-Networks} & 1500 & 2.25M\\ \hline
\end{tabular}
\centering
\label{dataset_summary}
\vspace*{-2mm}
\end{table}

\subsection{Data Preprocessing}
\label{subsec-data-preproc}

For each dataset, we randomly split 60\%, 20\%, and 20\% of all the graphs as training set, validation set, and testing set, respectively. The evaluation reflects the real-world scenario of graph query: For each graph in the testing set, we treat it as a query graph, and let the model compute the similarity between the query graph and every graph in the database. The database graphs are ranked according to the computed similarities to the query.

Since graphs from AIDS and LINUX are relatively small, an exponential-time exact GED computation algorithm named A*~\cite{riesen2013novel} is used to compute the GEDs between all the graph pairs. For the IMDB dataset, however, A* can no longer be used, as a recent survey of exact GED computation~\cite{blumenthal2018exact} concludes, ``no currently available algorithm manages to reliably compute GED within reasonable time between graphs with more than 16 nodes.'' 

To properly handle the IMDB dataset, we take the smallest distance computed by three well-known approximate algorithms, 	Beam~\cite{neuhaus2006fast}, Hungarian~\cite{kuhn1955hungarian,riesen2009approximate}, and VJ~\cite{jonker1987shortest,fankhauser2011speeding}. The minimum is taken instead of the average, because their returned GEDs are guaranteed to be greater than or equal to the true GEDs. Details on these algorithms can be found in Section~\ref{subsec-baselines}. Incidentally, the ICPR 2016 Graph Distance Contest~\footnote{\url{https://gdc2016.greyc.fr/}} also adopts this approach to obtaining ground-truth GEDs for large graphs.

To transform ground-truth GEDs into ground-truth similarity scores to train our model, we first normalize the GEDs according to \cite{qureshi2007graph}: $\mathrm{nGED}(\mathcal{G}_1,\mathcal{G}_2)=\frac{\mathrm{GED}(\mathcal{G}_1,\mathcal{G}_2)} {(|\mathcal{G}_1| + |\mathcal{G}_2|) / 2}$, where $|\mathcal{G}_i|$ denotes the number of nodes of $\mathcal{G}_i$.
We then adopt the exponential function $\lambda(x) = e^{-x}$ to transform the normalized GED into a similarity score in the range of $(0, 1]$. Notice that there is a one-to-one mapping between the GED and the similarity score.

\subsection{Baseline Methods}
\label{subsec-baselines}

Our baselines include two types of approaches, fast approximate GED computation algorithms and neural network based models. 

The first category of baselines includes three classic algorithms for GED computation. (1) \textit{A*-Beamsearch (Beam)}~\cite{neuhaus2006fast}. It is a variant of the A* algorithm in sub-exponential time. (2) \textit{Hungarian}~\cite{kuhn1955hungarian,riesen2009approximate} and (3) \textit{VJ}~\cite{jonker1987shortest,fankhauser2011speeding} are two cubic-time algorithms based on the Hungarian Algorithm for bipartite graph matching, and the algorithm of Volgenant and Jonker, respectively.

The second category of baselines includes seven models of the following neural network architectures. (1) \textit{SimpleMean} simply takes the unweighted average of all the node embeddings of a graph to generate its graph-level embedding. (2) \textit{HierarchicalMean} and (3) \textit{HierarchicalMax}~\cite{defferrard2016convolutional} are the original GCN architectures based on graph coarsening, which use the global mean or max pooling to generate a graph hierarchy. We use the implementation from the Github repository of the first author of GCN~\footnote{\url{https://github.com/mdeff/cnn_graph}}. The next four models apply the attention mechanism on nodes. (4) \textit{AttDegree} uses the natural log of the degree of a node as its attention weight, as described in Section~\ref{subsec-graph-emb}. (5) \textit{AttGlobalContext} and (6) \textit{AttLearnableGlobalContext (AttLearnableGC)} both utilize the global graph context to compute the attention weights, but the former does not apply the nonlinear transformation with learnable weights on the context, while the latter does. (7) \textit{SimGNN} is our full model that combines the best of Strategy 1 (AttLearnableGC) and Strategy 2 as described in Section~\ref{subsec-pnc}.

\subsection{Parameter Settings} 
\label{subsec-param-set}

For the model architecture, we set the number of GCN layers to 3, and use ReLU as the activation function. For the initial node representations, we adopt the one-hot encoding scheme for AIDS reflecting the node type, and the constant encoding scheme for LINUX and IMDB, since their nodes are unlabeled, as mentioned in Section~\ref{subsec-node-emb}. The output dimensions for the 1st, 2nd, and 3rd layer of GCN are 64, 32, and 16, respectively. For the NTN layer, we set $K$ to 16. For the pairwise node comparison strategy, we set the number of histogram bins to 16. We use 4 fully connected layers to reduce the dimension of the concatenated results from the NTN module, from 32 to 16, 16 to 8, 8 to 4, and 4 to 1.

We conduct all the experiments on a single machine with an Intel i7-6800K CPU and one Nvidia Titan GPU. As for training, we set the batch size to 128, use the Adam algorithm for optimization~\cite{kingma2014adam}, and fix the initial learning rate to 0.001. We set the number of iterations to 10000, and select the best model based on the lowest validation loss. 

\subsection{Evaluation Metrics}

The following metrics are used to evaluate all the models: \textit{Time.} The wall time needed for each model to compute the similarity score for a pair of graphs is collected. \textit{Mean Squared Error (mse).} The mean squared error measures the average squared difference between the computed similarities and the ground-truth similarities. 

We also adopt the following metrics to evaluate the ranking results. \textit{Spearman's Rank Correlation Coefficient ($\rho$)}~\cite{spearman1904proof} and \textit{Kendall's Rank Correlation Coefficient ($\tau$)}~\cite{kendall1938new} measure how well the predicted ranking results match the true ranking results. \textit{Precision at $k$ (p@$k$).} p@$k$ is computed by taking the intersection of the predicted top $k$ results and the ground-truth top $k$ results divided by $k$. Compared with p@$k$, $\rho$ and $\tau$ evaluates the global ranking results instead of focusing on the top $k$ results.

\subsection{Results}

\subsubsection{Effectiveness}

The effectiveness results on the three datasets can be found in Table~\ref{aids_results}, \ref{linux_results}, and \ref{imdb_results}. Our model, SimGNN, consistently achieves the best or the second best performance on all metrics across the three datasets. Within the neural network based methods, SimGNN consistently achieves the best results on all metrics. This suggests that our model can learn a good embedding function that generalizes to unseen test graphs.

Beam achieves the best precisions at 10 on AIDS and LINUX. We conjecture that it can be attributed to the imbalanced ground-truth GED distributions. As seen in Fig.~\ref{fig:ged_distribution}, for AIDS, the training pairs have GEDs mostly around 10, causing our model to train the very similar pairs less frequently than the dissimilar ones. For LINUX, the situation for SimGNN is better, since most GEDs concentrate in the range of [0, 10], the gap between the precisions at 10 of Beam and SimGNN become smaller.

It is noteworthy that among the neural network based models, AttDegree achieves relatively good results on IMDB, but not on AIDS or LINUX. It could be due to the unique ego-network structures commonly present in IMDB. As seen in Fig.~\ref{fig:search_demo_imdb}, the high-degree central node denotes the particular actor/actress himself/herself, focusing on which could be a reasonable heuristic. In contrast, AttLearnableGC adapts to the GED metric via a learnable global context, and consistently performs better than AttDegree. Combined with Strategy 2, SimGNN achieves even better performances.

\begin{figure}
    \centering
    {{\includegraphics[scale=0.28]{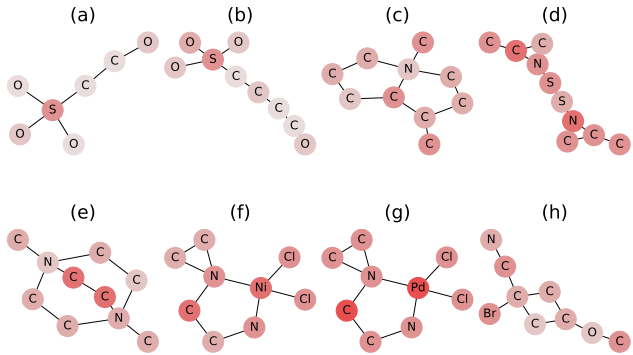} }}
    \vspace{-0.1in}
    \caption{Visualizations of node attentions. The darker the color, the larger the attention weight.
    }
    \vspace{-0.1in}
    \label{fig:att_vis}
\end{figure}

Visualizations of the node attentions can be seen in Fig.~\ref{fig:att_vis}. We observe that the following kinds of nodes receive relatively higher attention weights: hub nodes with large degrees, e.g. the ``S'' in (a) and (b), nodes with labels that rarely occur in the dataset, e.g. the ``Ni'' in (f), the ``Pd'' in (g), the ``Br'' in (h), nodes forming special substructures, e.g. the two middle ``C''s in (e), etc. These patterns make intuitive sense, further confirming the effectiveness of the proposed approach.

\begin{table}
\footnotesize
\centering
\vspace{-0.05in}
\caption{Results on AIDS.}
\vspace{-0.05in}
\begin{tabular}{|c|ccccc|} \hline
\textbf{Method} & \textbf{mse($10^{-3})$} & \textbf{$\rho$} & \textbf{$\tau$}&\textbf{p@10} & \textbf{p@20} \\ \hline\hline
\textbf{Beam} & $12.090$ & $0.609$ & $0.463$ & $\textbf{0.481}$ & $0.493$ \\
\textbf{Hungarian} & $25.296$ & $0.510$ & $0.378$& $0.360$ & $0.392$ \\
\textbf{VJ}  & $29.157$ & $0.517$ & $0.383$ &$0.310$ & $0.345$ \\ \hline\hline
\textbf{SimpleMean} & $3.115$ &  $0.633$ & $0.480$ & $0.269$ & $0.279$  \\
\textbf{HierarchicalMean} & $3.046$ &  $0.681$ & $0.629$ & $0.246$ & $0.340$ \\
\textbf{HierarchicalMax} & $3.396$ &  $0.655$ & $0.505$& $0.222$ & $0.295$ \\
\textbf{AttDegree} & $3.338$ & $0.628$ & $0.478$ & $0.209$ & $0.279$  \\
\textbf{AttGlobalContext} & $1.472$ &$0.813$ & $0.653$ & $0.376$ & $0.473$  \\
\textbf{AttLearnableGC} & $1.340$ & $0.825$ & $0.667$ &  $0.400$ & $0.488$  \\ \hline\hline
\textbf{SimGNN} & $\textbf{1.189}$ & $\textbf{0.843}$ & $\textbf{0.690}$ & $\textbf{0.421}$ & $\textbf{0.514}$ \\ \hline
\end{tabular}
\centering
\label{aids_results}
\end{table}

\begin{table}
\footnotesize
\centering
\vspace{-0.05in}
\caption{Results on LINUX.}
\vspace{-0.05in}
\begin{tabular}{|c|ccccc|} \hline
\textbf{Method} & \textbf{mse($10^{-3})$} & \textbf{$\rho$} & \textbf{$\tau$} &  \textbf{p@10} & \textbf{p@20} \\ \hline\hline
\textbf{Beam} & $9.268$ & $0.827$ & $0.714$  & $\textbf{0.973}$ & $0.924$ \\
\textbf{Hungarian} & $29.805$ & $0.638$ & $0.517$ & $0.913$ & $0.836$ \\
\textbf{VJ} & $63.863$ &  $0.581$ &  $0.450$ & $0.287$ & $0.251$ \\ \hline\hline
\textbf{SimpleMean} & $16.950$ & $0.020$ & $0.016$ & $0.432$ & $0.465$  \\
\textbf{HierarchicalMean} & $6.431$ &  $0.430$ & $0.525$ & $0.750$ & $0.618$ \\
\textbf{HierarchicalMax}  & $6.575$ & $0.879$ & $0.740$ & $0.551$ & $0.575$ \\
\textbf{AttDegree} & $8.064$ &  $0.742$ & $0.609$ & $0.427$ & $0.460$ \\
\textbf{AttGlobalContext} & $3.125$ & $0.904$ & $0.781$ & $0.874$ & $0.864$  \\
\textbf{AttLearnableGC} & $2.055$ & $0.916$ & $0.804$ & $0.903$ & $0.887$  \\ \hline\hline
\textbf{SimGNN} & $\textbf{1.509}$ & $\textbf{0.939}$ & $\textbf{0.830}$ & $\textbf{0.942}$ & $\textbf{0.933}$ \\ \hline
\end{tabular}
\centering
\label{linux_results}
\end{table}

\begin{table}
\footnotesize
\centering
\vspace{-0.05in}
\caption{Results on IMDB. Beam, Hungarian, and VJ together are used to determine the ground-truth results.}
\vspace{-0.05in}
\begin{tabular}{|c|ccccc|} \hline
\textbf{Method} & \textbf{mse($10^{-3})$} & \textbf{$\rho$} & \textbf{$\tau$}  & \textbf{p@10} & \textbf{p@20} \\ \hline\hline
\textbf{SimpleMean} & $3.749$ & $0.774$ & $0.644$ & $0.547$ & $0.588$  \\
\textbf{HierarchicalMean} & $5.019$ & $0.456$ & $0.378$ & $0.567$ & $0.553$  \\
\textbf{HierarchicalMax} & $6.993$ &$0.455$ & $0.354$ & $0.572$ & $0.570$  \\
\textbf{AttDegree} & $2.144$ & $0.828$ & $0.695$  & $0.700$ & $0.695$ \\
\textbf{AttGlobalContext} & $3.555$ & $0.684$ & $0.553$ & $0.657$ & $0.656$ \\
\textbf{AttLearnableGC} & $1.455$ & $0.835$ & $0.700$ & $0.732$ & $0.742$ \\ \hline\hline
\textbf{SimGNN} & $\textbf{1.264}$ &  $\textbf{0.878}$ & $\textbf{0.770}$ & $\textbf{0.759}$ & $\textbf{0.777}$ \\ \hline
\end{tabular}
\centering
\label{imdb_results}
\end{table}


\subsubsection{Efficiency}
\label{subsec-efficiency}
\begin{figure}
\centering
\includegraphics[scale=0.15]{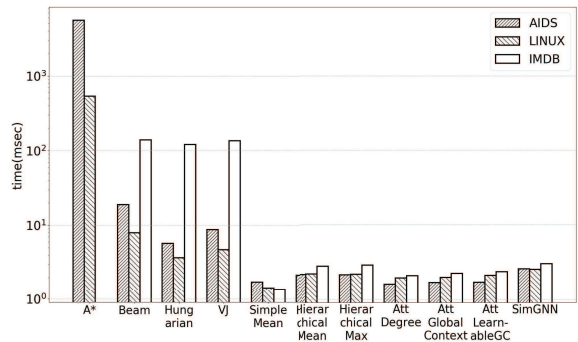}
\vspace{-0.05in}
\caption{Runtime comparison.}
\label{fig:time}
\vspace{-0.05in}
\end{figure}

The efficiency comparison on the three datasets is shown in Fig.~\ref{fig:time}. The neural network based models consistently achieve the best results across all the three datasets. Specifically, compared with the exact algorithm, A*, SimGNN is 2174 times faster on AIDS, and 212 times faster on LINUX. The A* algorithm cannot even be applied on large graphs, and in the case of IMDB, its variant, Beam, is still 46 times slower than SimGNN. Moreover, the time measured for SimGNN includes the time for graph embedding. As mentioned in Section~\ref{subsec-time-comp-analysis}, if graph embeddings are pre-computed and stored, SimGNN would spend even less time. All of these suggest that in practice, it is reasonable to use SimGNN as a fast approach to graph similarity computation, which is especially true for large graphs, as in IMDB, our computation time does not increase much compared with AIDS and LINUX.

\subsection{Parameter Sensitivity}

\begin{figure}
    \centering
    \subfloat[]{{\includegraphics[scale=0.22]{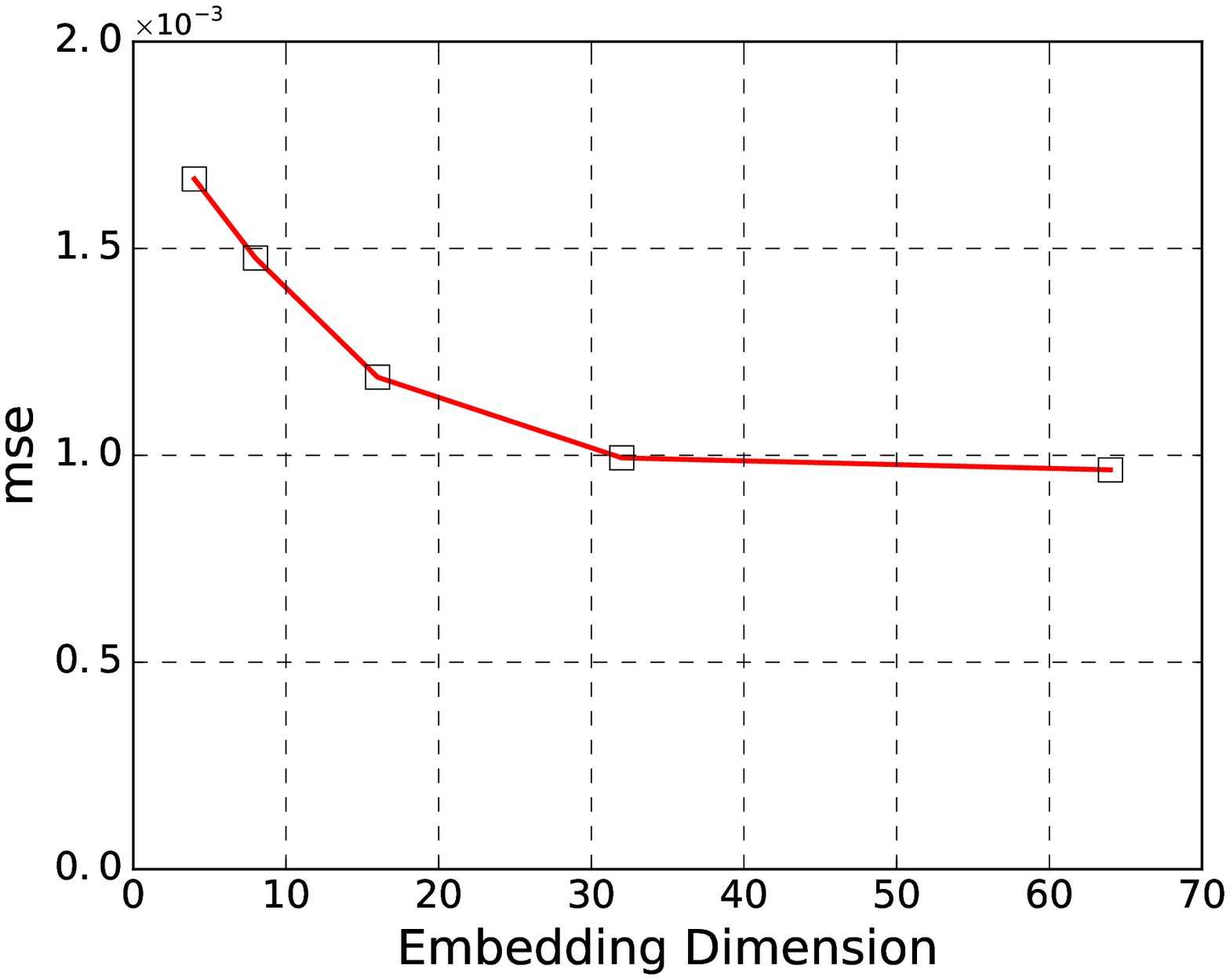} } \label{fig:param_emb}}
    \subfloat[]{{\includegraphics[scale=0.22]{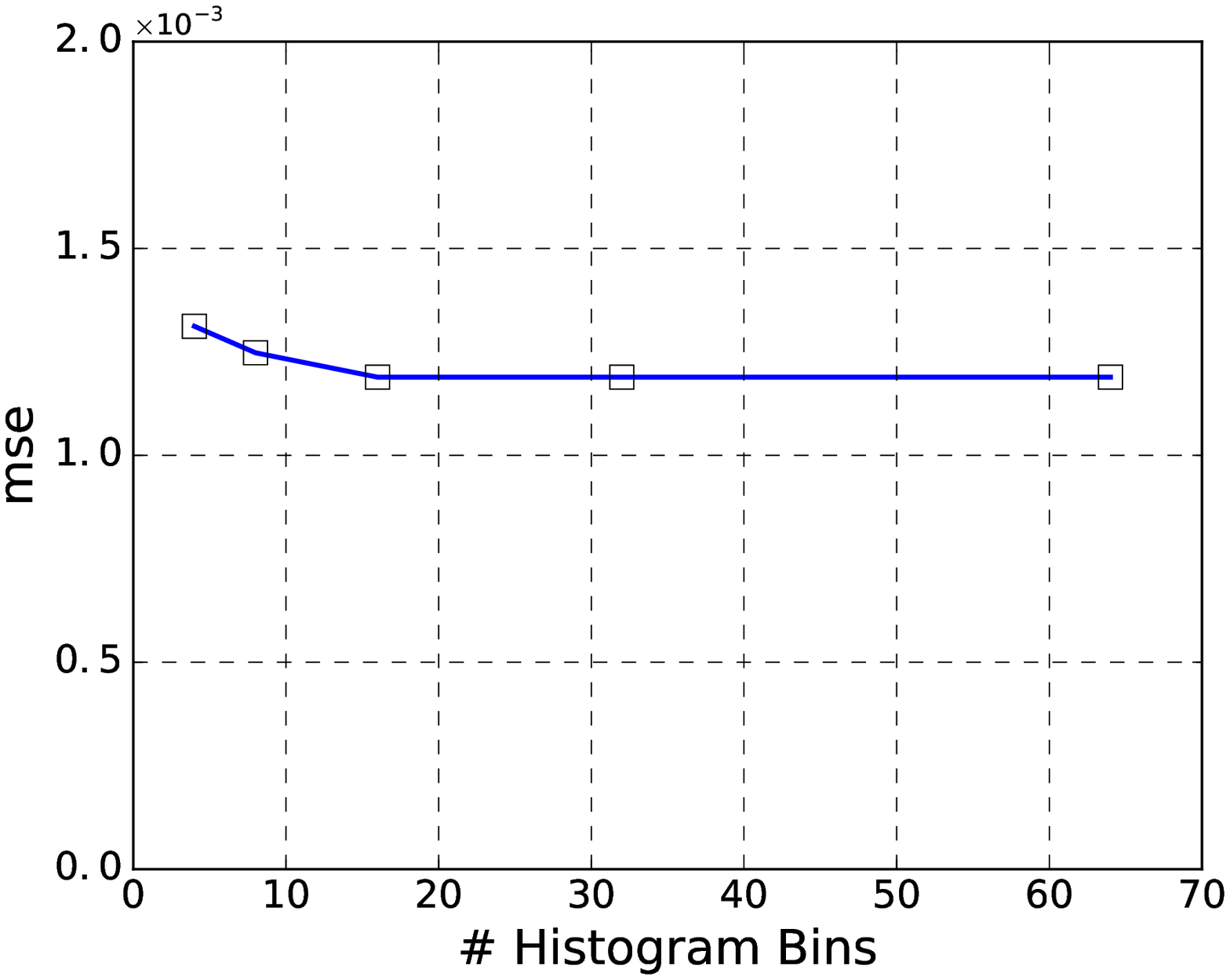} \label{fig:param_bin}}}
      \vspace{-0.1in}
    \caption{Mean squared error w.r.t. the number of dimensions of graph-level embeddings, and the number of histogram bins.}
    \vspace{-0.05in}
    \label{fig:param}
\end{figure}

We evaluate how the dimension of the graph-level embeddings and the number of the histogram bins can affect the results. We report the mean squared error on AIDS. As can be seen in Fig.~\ref{fig:param_emb}, the performance becomes better if larger dimensions are used. This makes intuitive sense, since larger embedding dimensions give the model more capacity to represent graphs. In our Strategy 2, as shown in Fig.~\ref{fig:param_bin}, the performance is relatively insensitive to the number of histogram bins. This suggests that in practice, as long as the histogram bins are not too few, relatively good performance can be achieved.

\subsection{Case Studies}

\begin{figure}
    \centering
    {{\includegraphics[scale=0.4]{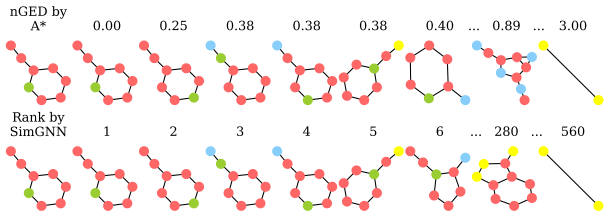} }}
    \vspace{-0.1in}
    \caption{A query case study on AIDS. Meanings of the colors can be found in Fig.~\ref{fig:aids_pie}. 
    }
    \vspace{-0.1in}
    \label{fig:search_demo_aids}
\end{figure}

\begin{figure}
    \centering
    {{\includegraphics[scale=0.4]{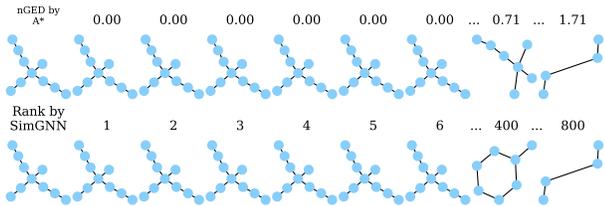} }}
\vspace{-0.1in}
    \caption{A query case study on LINUX.}
    \vspace{-0.1in}
    \label{fig:search_demo_linux}
\end{figure}

\begin{figure}
    \centering
    {{\includegraphics[scale=0.4]{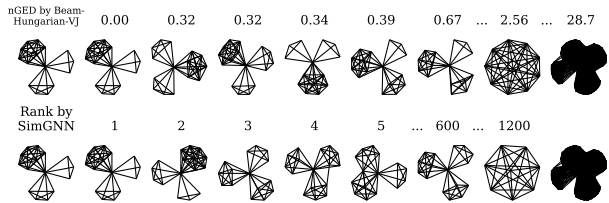} }}
\vspace{-0.1in}
    \caption{A query case study on IMDB.}
    \vspace{-0.1in}
    \label{fig:search_demo_imdb}
\end{figure}

We demonstrate three example queries, one from each dataset, in Fig.~\ref{fig:search_demo_aids}, \ref{fig:search_demo_linux}, and \ref{fig:search_demo_imdb}. In each demo, the top row depicts the query along with the ground-truth ranking results, labeled with their normalized GEDs to the query; The bottom row shows the graphs returned by our model, each with its rank shown at the top. SimGNN is able to retrieve graphs similar to the query, e.g. in the case of LINUX (Fig.~\ref{fig:search_demo_linux}), the top 6 results are exactly the isomorphic graphs to the query.

\section{Related Work}
\label{sec-related}

To precisely position our task, outline the scope of our work, and compare various methods related to our model, we briefly survey the following topics.
\subsection{Network/Graph Embedding}
\textbf{Node-level embedding}. Over the years, there are several categories of methods that have been proposed for learning node representations, including matrix factorization based methods (NetMF~\cite{qiu2018network}), skip-gram based methods (DeepWalk~\cite{perozzi2014deepwalk}, Node2Vec~\cite{grover2016node2vec}, LINE~\cite{tang2015line}), autoencoder based methods (SDNE~\cite{wang2016structural}), neighbor aggregation based methods (GCN~\cite{defferrard2016convolutional,kipf2016semi,kipf2016variational}, GraphSAGE~\cite{hamilton2017inductive}), etc. 

\noindent\textbf{Graph-level embedding}. The most intuitive way to generate one embedding per graph is to aggregate the node-level embeddings, either by a simple average or some weighted average~\cite{duvenaud2015convolutional,dai2016discriminative,zhao2018substructure}, named the ``sum-based'' approaches~\cite{hamilton2017representation}.  A more sophisticated way to represent graphs can be achieved by viewing a graph as a hierarchical data structure and applying graph coarsening~\cite{bruna2013spectral,defferrard2016convolutional,simonovsky2017dynamic,ying2018hierarchical}. Besides, \cite{kearnes2016molecular} aggregate sets of nodes via histograms, and \cite{niepert2016learning} applies node ordering on a graph to make it CNN suitable.

\noindent\textbf{Graph neural network applications}. A great amount of graph-based applications have been tackled by neural network based methods, most of which are framed as node-level prediction tasks.
However, once moving to the graph-level tasks, most existing work deal with the classification of a single graph~\cite{defferrard2016convolutional,simonovsky2017dynamic,niepert2016learning,simonovsky2017dynamic,gligorijevic2017deepnf,zhao2018substructure,ying2018hierarchical}. 
In this work, we consider the task of graph similarity computation for the first time.


\subsection{Graph Similarity Computation}
\noindent\textbf{Graph distance/similarity metrics.} The Graph Edit Distance (GED)~\cite{bunke1983distance} can be considered as an extension of the String Edit Distance metric~\cite{levenshtein1966binary}, which is defined as the minimum cost taken to transform one graph to the other via a sequence graph edit operations. Another well-known concept is the Maximum Common Subgraph (MCS), which has been shown to be equivalent to GED under a certain cost function~\cite{bunke1997relation}. Graph kernels can be considered as a family of different graph similarity metrics, used primarily for graph classification. Numerous graph kernels~\cite{gartner2003graph,horvath2004cyclic,nikolentzos2017matching} and extensions~\cite{yanardag2015deep,nikolentzos2018degeneracy} have been proposed across the years.

\noindent\textbf{Pairwise GED computation algorithms}. 
A flurry of approximate algorithms has been proposed to reduce the time complexity with the sacrifice in accuracy \cite{neuhaus2006fast,riesen2009approximate,fankhauser2011speeding,bougleux2017graph,daller2018approximate}.  We are aware of some very recent work claiming their time complexity is $O(n^2)$~\cite{bougleux2017graph}, but their code is unstable at this stage for comparison.

\noindent\textbf{Graph Similarity search}. 
Computing GED is a primitive operator in graph database analysis, and has been adopted in a series of works on graph similarity search~\cite{zeng2009comparing,wang2012efficient,zheng2013graph,zhao2013partition,liang2017similarity}. It must be noted, however, that these studies focus on database-level techniques to speed up the overall querying process involving exact GED computations, while our model, at the current stage, is more comparable in its flavor to the approximate pairwise GED computation algorithms.


\section{Discussions and Future Directions}
\label{sec-future}
There are several directions to go for the future work: (1) our model can handle graphs with node types but cannot process edge features. In chemistry, bonds of a chemical compound are usually labeled, so it is useful to incorporate edge labels into our model;
(2) it is promising to explore different techniques to further boost the precisions at the top k results, which is not preserved well mainly due to the skewed similarity distribution in the training dataset; and (3) given the constraint that the exact GEDs for large graphs cannot be computed, it would be interesting to see how the learned model generalize to large graphs, which is trained only on the exact GEDs between small graphs.
\section{Conclusion}
We are at the intersection of graph deep learning and graph search problem, and taking the first step towards bridging the gap, by tackling the core operation of graph similarity computation
, via a novel neural network based approach. The central idea is to learn a neural network based function that is representation-invariant, inductive, and adaptive to the specific similarity metric, which takes any two graphs as input and outputs their similarity score. 
Our model runs very fast compared to existing classic algorithms on approximate Graph Edit Distance computation, and achieves very competitive accuracy.
\label{sec-conc}

\subsection*{Acknowledgments}

The work is supported in part by NSF DBI 1565137, NSF DGE1829071, NSF III-1705169, NSF CAREER Award 1741634, NIH U01HG008488, NIH R01GM115833, Snapchat gift funds, and PPDai gift fund.

\bibliographystyle{ACM-Reference-Format}
\bibliography{bibliography}

\end{document}